\documentclass[conference]{IEEEtran}
\IEEEoverridecommandlockouts
\usepackage{cite}
\usepackage{amsmath,amssymb,amsfonts}
\usepackage{algorithmic}
\usepackage{graphicx}
\usepackage{xcolor}
\usepackage{textcomp}
\usepackage{colortbl}
\usepackage[numbers,sort&compress]{natbib}
\usepackage{hyperref}
\usepackage{booktabs}
\usepackage{multirow}
\usepackage{pifont}

\def\BibTeX{{\rm B\kern-.05em{\sc i\kern-.025em b}\kern-.08em
    T\kern-.1667em\lower.7ex\hbox{E}\kern-.125emX}}

\newcommand{\method}{\textcolor{black}{{\texttt{PosePilot}}}}

\begin{document}

\title{PosePilot: Steering Camera Pose for Generative World Models with Self-supervised Depth} 

\author{Bu Jin$^{1,3*}$, Weize Li$^{2*}$, Baihan Yang$^{2}$, Zhenxin Zhu$^{2}$, Junpeng Jiang$^{3}$, Huan-ang Gao$^{2}$, 
Haiyang Sun$^{3}$\\ Kun Zhan$^{3}$, Hengtong Hu$^{3}$, Xueyang Zhang$^{3}$, Peng Jia$^{3}$, Hao Zhao$^{2,4\dagger}$

\thanks{$^{*}$ indicates equal contribution and $^{\dagger}$ indicates the corresponding author.}
\thanks{$^{1}$The Hong Kong University of Science and Technology.}
\thanks{$^{2}$Institute for AI Industry Research (AIR), Tsinghua University.}
\thanks{$^{3}$Li Auto.  $^{4}$BAAI.} 
% \thanks{}
}

\maketitle
\begin{abstract}
Recent advancements in autonomous driving (AD) systems have highlighted the potential of world models in achieving robust and generalizable performance across both ordinary and challenging driving conditions. However, a key challenge remains: precise and flexible camera pose control, which is crucial for accurate viewpoint transformation and realistic simulation of scene dynamics. In this paper, we introduce PosePilot, a lightweight yet powerful framework that significantly enhances camera pose controllability in generative world models. Drawing inspiration from self-supervised depth estimation, PosePilot leverages structure-from-motion principles to establish a tight coupling between camera pose and video generation. Specifically, we incorporate self-supervised depth and pose readouts, allowing the model to infer depth and relative camera motion directly from video sequences. These outputs drive pose-aware frame warping, guided by a photometric warping loss that enforces geometric consistency across synthesized frames. To further refine camera pose estimation, we introduce a reverse warping step and a pose regression loss, improving viewpoint precision and adaptability. Extensive experiments on autonomous driving and general-domain video datasets demonstrate that PosePilot significantly enhances structural understanding and motion reasoning in both diffusion-based and auto-regressive world models. By steering camera pose with self-supervised depth, PosePilot sets a new benchmark for pose controllability, enabling physically consistent, reliable viewpoint synthesis in generative world models.
\end{abstract}


\begin{figure*}[!h]
    \centering
    \includegraphics[width=\textwidth]{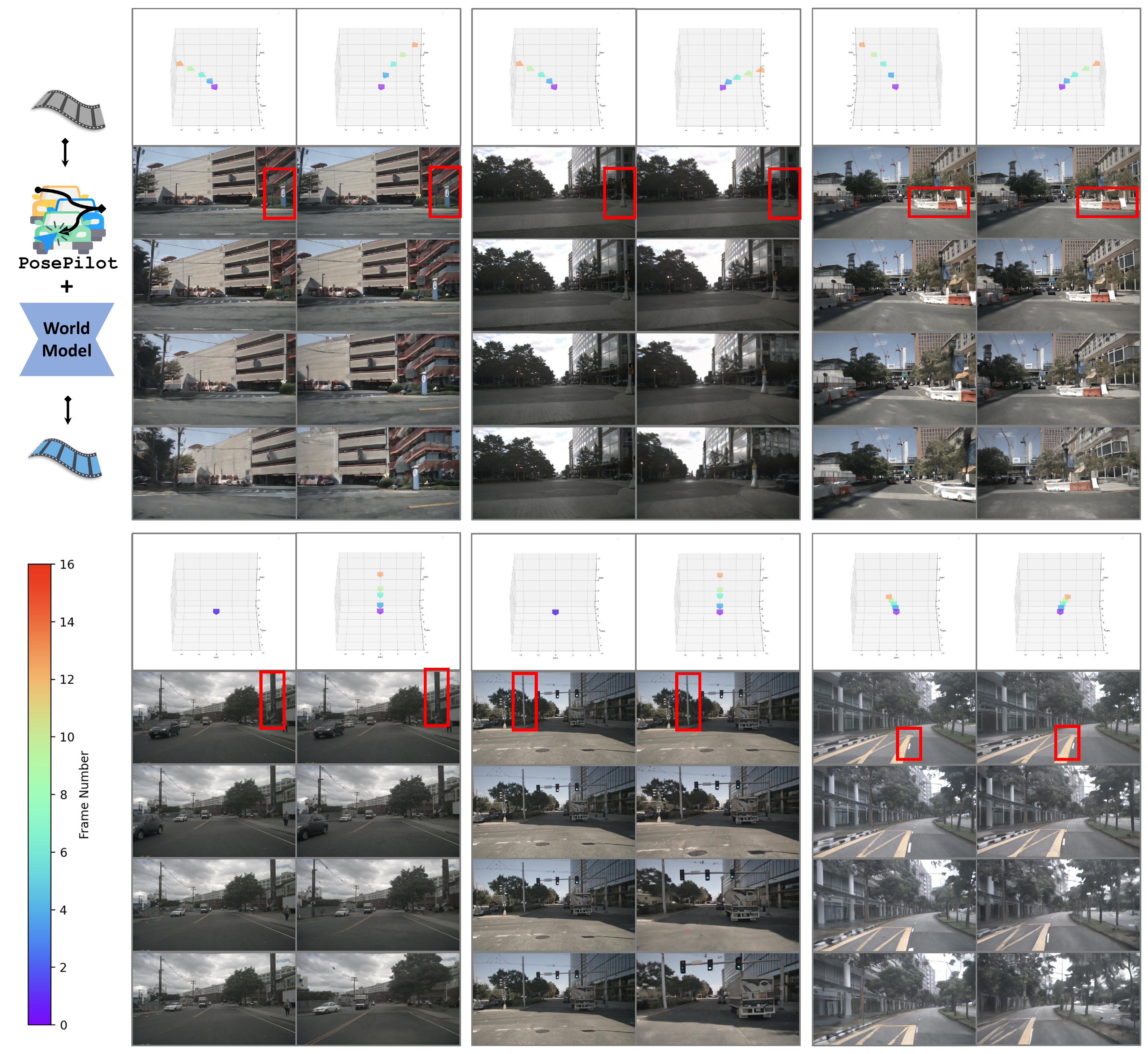}
    \vspace{-20pt}
    \caption{\textbf{Illustration of \method}. We propose \method, a lightweight pose controllability enhancer for world models in autonomous driving. \method can generate videos collaboratively aligned with the input camera pose. The \textcolor{red}{red box} indicates the reference objects for easy identification.}
    \label{fig:teaser}
\end{figure*}

\section{Introduction}
In recent years, great advancements have been made in the development of autonomous driving systems. While existing AD methods \cite{hu2023planning, yang2024unipad,jiang2024p,li2023lode,zheng2024monoocc,tian2023unsupervised,jin2024tod3cap,li2025uniscene,ding2024hint} demonstrate promising results in various scenarios, they still face significant challenges when confronted with long-tail distribution or out-of-distribution situations. 
These edge cases, which occur infrequently in typical training datasets but are critical in real-world driving environments, expose limitations in the robustness and generalization ability of current models. 
One promising solution is world models \cite{hu2022MILE, drivingintothefuture, wang2023drivedreamer, bevcontrol,gao2023magicdrive, panacea, huang2024subjectdrive, gao2024vista}, which capture the structure and dynamics of an environment, enabling an autonomous driving system to simulate and predict future states by "imagining" the external world. Through building such models, these generative models can help AD systems anticipate future states, reason about complex dynamics and better generalize to novel or unexpected situations.

A critical factor in harnessing the full potential of world models lies in \textbf{pose controllability}, the ability to manipulate and reason about an agent’s location, orientation, and viewpoint within the learned representation. Pose controllability ensures that the model can adapt its predictive capabilities to varying viewpoints, accurately simulate sensor feedback, and better handle environmental uncertainties. By granting precise and flexible control over the pose, the world model gains a robust mechanism for exploring and understanding the state space, thereby facilitating improved performance in both typical and hard-to-generalize driving conditions.

Existing works \cite{gao2024vista, wang2024motionctrl, yang2024direct, he2024cameractrl} have tried to introduce the pose controllability to the video generation model. For example, AnimateDiff \cite{guo2023animatediff} introduces a transformer-based motion module along with a MotionLoRA, supporting certain types of camera movement. MotionCtrl \cite{wang2024motionctrl} encodes the camera pose values $RT$ with several fully connected layers and inserts the embeddings to the transformer layers to interact with the visual outputs, enabling a flexible camera pose control. CameraCtrl \cite{he2024cameractrl} further utilizes a plücker embeddings as the primary form of camera parameters. Vista \cite{gao2024vista} proposes a versatile action controllability with diverse control formats, like motion values, camera trajectory, driving command or goal point.
However, these works typically learn the controllability through learnable attention layers, lacking a deeper understanding of the relation between camera pose and the structural evolution of the scene.

Inspired from self-supervised depth estimation \cite{zhou2017unsupervised, mahjourian2018unsupervised, bian2019unsupervised, garg2016unsupervised, bian2021unsupervised}, the camera pose is intrinsically tied to adjacent frames in a video. Specifically, the transformation between two neighboring frames, parameterized by camera intrinsics and their relative poses, enables warping pixels from one image onto another through the corresponding depth map. This structure-from-motion property ensures geometric consistency across different camera motions and provides an effective prior for pose controllability in generative world models.

We thus propose \method, a light-weight camera pose control method that enhances the pose controllability in world models.
Specifically, we introduce a depth readout and an ego‑motion readout to generate the depth maps and camera poses of the input video. These outputs are then used to warp a source frame onto its neighboring frame with a specified camera pose. The warped frames guide the synthesized video through a photometric control loss \cite{garg2016unsupervised}, ensuring that the generated video remains consistent with the specified camera pose. By enforcing this warping consistency, \method~ enhances the world model's ability to reason about its location, orientation, and viewpoint, thereby improving pose controllability.
Furthermore, we introduce an inverse photometric control loss, which applies an inverse warping step to the synthesized video, aligning it with the original video via an additional pair of depth and pose readouts to further enhance the controllability. We also introduce a regression loss to the pose readouts, explicitly refining the pose estimation to achieve more precise pose control. This multi‑stage procedure bolsters the overall pose controllability of the world models, ensuring more accurate and reliable camera viewpoints. Our contributions can be summarized as follows:
\vspace{-1pt}
\begin{itemize}
    \item We propose \method, a lightweight, plug-and-play module that injects camera pose controllability into generative world models. It supports flexible viewpoint control across both diffusion and autoregressive architectures.
    \item We design a photometric warping supervision strategy that leverages self-supervised depth and ego-motion estimation to enforce forward and inverse geometric consistency in video generation.
    \item Extensive experiments and ablations on driving and real-world datasets show that \method~improves pose accuracy, structural coherence, and generalization across diverse generative models.
\end{itemize}

\begin{figure*}[!h]
    \centering
    \includegraphics[width=0.9\textwidth]{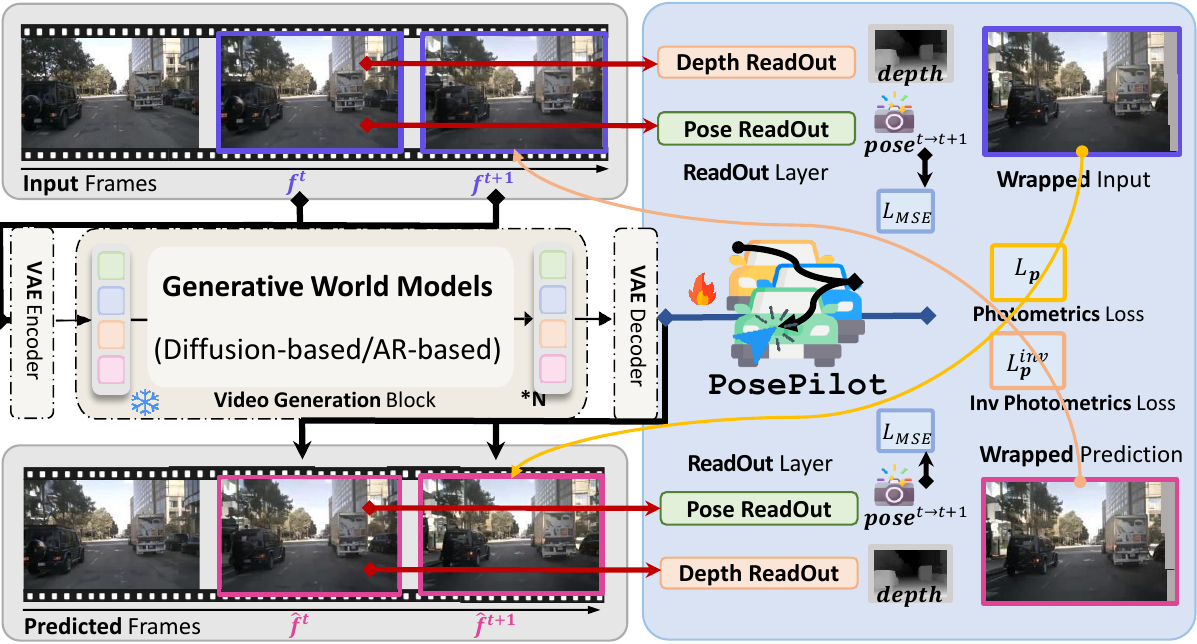}
    \vspace{-2pt}
    \caption{\textbf{Overview of \method.} We utilize off-the-shelf generative world model as our base model and build our \method~ on top of them. We introduce two auxiliary tasks: depth estimation and ego-motion estimation. The depth estimation predicts per-pixel depth maps, while ego-motion prediction estimates the relative camera pose between consecutive frames. We introduce a photometric control loss and an inverse photometric control loss, which aligns the predicted frames with real frames by warping pixels based on the estimated depth and ego-motion.}
    \label{fig:main}
\end{figure*}

\section{Related Works}
\subsection{World Models for Driving Scenes}
Generative world models World Models~\cite{feng2025survey,chen2024end,gu2024dome,zheng2025world4drive} have emerged as a promising paradigm for simulating complex driving environments by predicting future visual states. Recent advances enabling video generation to augment autonomous driving data with diverse road conditions and rare corner cases~\cite{yan2023int2,bai2022transfusion,ding2023pivotnet}.
Existing methods use various conditions including images, text, 3D layouts, and actions to generate specific driving scenes or predict future video sequences.
GAIA-1~\cite{hu2023gaia} generates realistic and diverse driving videos by integrating video, text, and action inputs, demonstrating a strong understanding of contextual information and physical principles.
DriveDreamer~\cite{wang2024drivedreamer} improves driving scenario generation by adding HD maps and 3D boxes for better video quality and enabling the generation of future driving actions, aiding in decision-making.
WorldDreamer~\cite{wang2024worlddreamer} frames world modeling as an unsupervised challenge within visual sequence modeling, inspired by large language models.
MUVO~\cite{bogdoll2023muvo} enhances world modeling by integrating LIDAR point clouds, improving the prediction of driving environments and generating 3D occupancy grids that integrate well with downstream tasks.
OccWorld~\cite{zheng2024occworld} leverages 3D occupancy data to predict environmental evolution and guide autonomous vehicle actions, marking a shift toward multimodal approaches in autonomous driving research.
In contrast, our work does not design a brand new framework for world model but serves as a plug-and-play module that builds upon existing world models to provide effective camera pose control during video generation for driving scenes. 

\subsection{Controllable Video Generation}
% What is controllable video generation, and we mainly care about camera pose control.
Controllability plays a crucial role in video generation applications~\cite{wu2025motionbooth, hou2024training, popov2025camctrl3d, zheng2024cami2v, feng2024i2vcontrol}, especially in enabling customized camera pose generation to better meet user needs.
%list previous camera ctrl works in general video generation.
MotionCtrl~\cite{wang2024motionctrl} enables independent control of camera and object motion in video generation by explicitly modeling camera poses and trajectories.
CameraCtrl~\cite{he2024cameractrl} enhances text-to-video generation with precise camera pose control using a plug-and-play module trained on parameterized trajectories.
CamTrol~\cite{hou2024training} achieves camera pose control in video diffusion models by leveraging latent layout priors to adjust noisy latents without fine-tuning.
CamCtrl3D\cite{popov2025camctrl3d} improves fly-through video generation by conditioning an image-to-video diffusion model on camera trajectories using multiple techniques.
In particular, controllable video generation in autonomous driving places special emphasis on camera pose control:
MagicDrive~\cite{gao2023magicdrive} enables camera pose-controllable street view generation by conditioning on 3D geometry inputs and ensuring cross-view consistency.
Vista~\cite{gao2024vista} enhances autonomous driving video generation with explicit motion dynamics and structural learning for high-fidelity, generalizable viewpoint control.
DiVE\cite{jiang2024dive} achieves multi-view consistent video generation by enforcing BEV layout control and integrating spatial attention mechanisms for precise camera pose alignment.

What sets \method~apart is that we explicitly read out depth and camera pose between adjacent frames for direct warping of future frames, rather than relying on other modalities as conditions, which can lead to suboptimal optimization gaps.

\section{Methodology}
We propose \method, a lightweight and modular framework designed to enhance pose controllability in generative world models. The core idea as shown in Fig.~\ref{fig:main} is to inject self-supervised geometric supervision into the video generation process, enabling the model to reason about camera motion in a structurally grounded way.
% \subsection{Preliminary}
We adapt \method~on top of the powerful generative world models \cite{jiang2024dive, gao2024vista, hu2024drivingworld}. 

\subsection{Pose-aware Warping via Self-supervised Geometry}
In contrast to traditional video generation models, a key factor in the world models for autonomous driving is the camera parameters. Once the extrinsic matrices (or pose) of each camera are avaiGiven two framesle, we can use the structure-from-motion methods to generate the structural and dynamic context to guide the video generation. Inspired by monocular self-supervised depth estimation \cite{bian2019unsupervised}, we incorporate a photometric control loss into our generative model. The objective of the photometric control loss is to derive real-world scale information from the camera's extrinsic matrices, which provides prior information for video generation.

Firstly, we introduce a learnable depth readout and a learnable pose readout, which generates the depth maps and relative 6D camera pose for two consecutive frames. Given two frames $(f^i, f^j)$, the outputs of the depth readout are their depth maps $(D^i, D^j)$, and the output of the pose readout is the relative 6D camera pose $T^{i \rightarrow j}$. These two values can subsequently be employed to warp the initial frame $f^i$ to $f^j$. Let $x^{i}$ be the coordinates of a pixel in the frame $f^i$, the projected coordinates  $x^{i \rightarrow j}$ in warped frame $f^{i\rightarrow j}$ can be calculated by:
\begin{equation}
x^{i \rightarrow j}=KT^{i \rightarrow j} D^i K^{-1} x^i,
\end{equation}
where $K$ are intrinsic matrices of the camera. The whole process is differentiable, following common practice in self-supervised depth estimation \cite{zhou2017unsupervised}.

With the synthesized $f'^j$ and the reference frame $\hat{f}^j$, the photometric control loss can be formulated as:
\begin{equation}
L_p=\frac{1}{|N|} \sum_{x \in N}\left\|\hat{f}^j(x)-f^{i\rightarrow j}(x)\right\|_1,
\end{equation}
where $\hat{f}$ represents the predicted frame generated by the generation model and $N$ denotes the collection of valid points mapped from $f^i$ onto the plane of $f^j$. $|N|$ represents the point number of $N$. Note that to better integrate the camera pose prior to guide the models' generation, our reference frames are not the input frames $f^j$, but the frames generated by the world models $\hat{f}^j$. In this way, the generation model can get a feedback from the photometric control loss, which enhances the controllability of camera pose.

To better handle the complex illumination changes, we also add auxiliary SSIM \cite{wang2004image} loss to the photometric control loss:
\begin{equation}
\begin{aligned}
L_p= & \frac{1}{|N|} \sum_{x \in N}(\left\|\hat{f}^j(x)-f^{i\rightarrow j}(x)\right\|_1 \\ 
& +\frac{1-\operatorname{SSIM}_{\hat{j}j'}(x)}{2}),
\end{aligned}
\end{equation}

where $\operatorname{SSIM}_{jj'}$ represents the per-element structural similarity between $\hat{f}^j$ and $f'^j$.

Through this photometric loss, our model is capable of capturing structural information from the vehicle's motion, which is essential for pose controllability for world models.

\subsection{Bidirectional Consistency for Inverse Warping}
Forward-only supervision may suffer from occlusions and partial observability. To reinforce geometric alignment, we introduce inverse photometric warping, which enforces consistency from generated frames back to the original viewpoints. This bidirectional alignment helps reduce the artifacts due to occlusions and dis-occlusions.

We denote the generated frames as $(\hat{f}^i, \hat{f}^j)$, which are obtained from the world model. The learnable depth readout provides their respective depth maps $(D^i, D^j)$, and the pose readout outputs the relative 6D camera transformations $T^{i \rightarrow j}$ and $T^{j \rightarrow i}$. To map $\hat{f}^j$ back to $\hat{f}^i$, we warp usxing the same bilinear sampling mechanism as above. Specifically, the pixel projection from $\hat{f}^j$ to the warped inverse view $\hat{f}^{j \rightarrow i}$ follows the same procedure as in the forward warping. Denoting $x^j$ as a pixel in the generated frame $\hat{f}^j$, its inverse-warped coordinate $x^{j \rightarrow i}$ is found by:
\begin{equation}
x^{j \rightarrow i} = KT^{j \rightarrow i}D^j K^{-1}p^j,
\end{equation}
where $K$ is the intrinsic matrix, and $D^j$ is the depth at pixel $x^j$ in the generated frame $\hat{f}^j$. 
Just as in the forward direction, this process is fully differentiable. Once we obtain the warped image $\hat{f}^{j \rightarrow i}$, we compute the inverse photometric control loss between $\hat{f}^i$ and $\hat{f}^{j \rightarrow i}$. Let $\tilde{N}$ denote the set of valid pixels that can be projected from $\hat{f}^j$ to $\hat{f}^i$. 
The inverse photometric control loss $L_p^{\text{inv}}$ has the same form as the photometric control loss, which includes the per-pixel L1 term and the auxiliary SSIM loss.
By enforcing consistency in this reverse mapping, the model refines the pose and depth estimation for generated frames and reduces warping artifacts in the generated views. This inverse photometric control loss, combined with the forward photometric control loss, ensures tighter motion consistency for video generation. 

\subsection{Refining Pose Estimates with Regression Supervision}
To further improve pose precision, we apply direct supervision to the pose readout module using a mean squared error (MSE) loss. Given a reference pose $T_{ref}$, the predicted pose $\hat{T}$ is refined via:
\begin{equation}
L_{\text{MSE}} = \left\| \hat{T} - T_{\text{ref}} \right\|_2^2
\end{equation}
This loss reduces drift in relative pose estimation and provides stable gradients for training under large motion variation.
% To further improve the precision of pose estimation, we introduce a pose regression loss that supervises the output of the pose readout module directly against ground-truth or synthetic reference poses: $L_{MSE}$. This term helps eliminate accumulated drift in relative pose prediction and provides a numerically stable learning signal for training on diverse trajectories.

\subsection{Overall Objective}
We combine the generative loss $L_{g}$ (e.g., diffusion or autoregressive objective) with the three geometric consistency terms into the final objective:
\begin{equation}
L_{\text{total}} = L_{g} + \alpha_p\,L_p + \alpha_{p^\text{inv}}\,L_p^{\text{inv}} + \alpha_{mse} L_{MSE},
\end{equation}
Here, $\alpha_p$, $\alpha_{p^\text{inv}}$ and $\alpha_{mse}$ are weights that balance fidelity and geometric alignment. In practice, these terms guide the generator to synthesize high-quality, structurally plausible videos that accurately reflect the specified camera trajectory.

% We combine the forward and inverse photometric control losses with our generative objective into a comprehensive total loss. 
% Let $L_{g}$ represent the generative objective (e.g., the diffusion-based reconstruction or adversarial loss), $L_p$ be the forward photometric control loss, and $L_p^{\text{inv}}$ be the inverse photometric control loss. We employ weighting factors $\alpha_p$ and $\alpha_{p^\text{inv}}$ to balance the relative importance of these terms, which yields the total loss:
% \begin{equation}
% L_{\text{total}} = L_{g} + \alpha_p\,L_p + \alpha_{p^\text{inv}}\,L_p^{\text{inv}} + \alpha_{mse} L_{MSE},
% \end{equation}
% where the $L_{MSE}$ represents the regression loss to the pose readouts and $\alpha_{mse}$ is its factor, explicitly refining the pose estimation to achieve more precise pose control.
% In practice, $\alpha_p$, $\alpha_{p^\text{inv}}$ and $\alpha_{mse}$ can be tuned to find an optimal trade-off between the generative fidelity of the generative model and the geometric consistency enforced by both the forward and inverse photometric control terms. 
% By minimizing $L_{\text{total}}$, the framework simultaneously learns to produce high-quality, realistic frames while ensuring that the predicted depth and pose remain consistent in both directions of warping.

\begin{table}[]
\begin{center} 
\caption{Comparison with previous works}
\label{tab:main}
\begin{tabular}{@{}lcc@{}} 
\toprule
\textbf{Method} & \textbf{\texttt{TransErr}}↓  & \textbf{\texttt{RotErr}}↓ \\
\midrule
\rowcolor{lightgray}\textbf{Diffusion-based Generation:} & &\\
DiVE \cite{jiang2024dive} & 13.07 & 4.52 \\
DiVE + \method~\textbf{(Ours)} & \textbf{6.37} & \textbf{1.40}\\
Vista~\cite{gao2024vista} & 6.83 & 1.74 \\
Vista + \method~\textbf{(Ours)} & \textbf{6.52} & \textbf{1.53}\\
\midrule
\rowcolor{lightgray}\textbf{Autoregressive-based Generation:} & &\\
DrivingWorld~\cite{hu2024drivingworld} & 3.17 & 1.64\\
DrivingWorld + \method~\textbf{(Ours)} & \textbf{2.95} & \textbf{1.48}\\
\bottomrule
\end{tabular}
\end{center}
\vspace{-0.5cm}
\end{table}

\begin{figure*}[!h]
    \centering
    \includegraphics[width=1.0\textwidth]{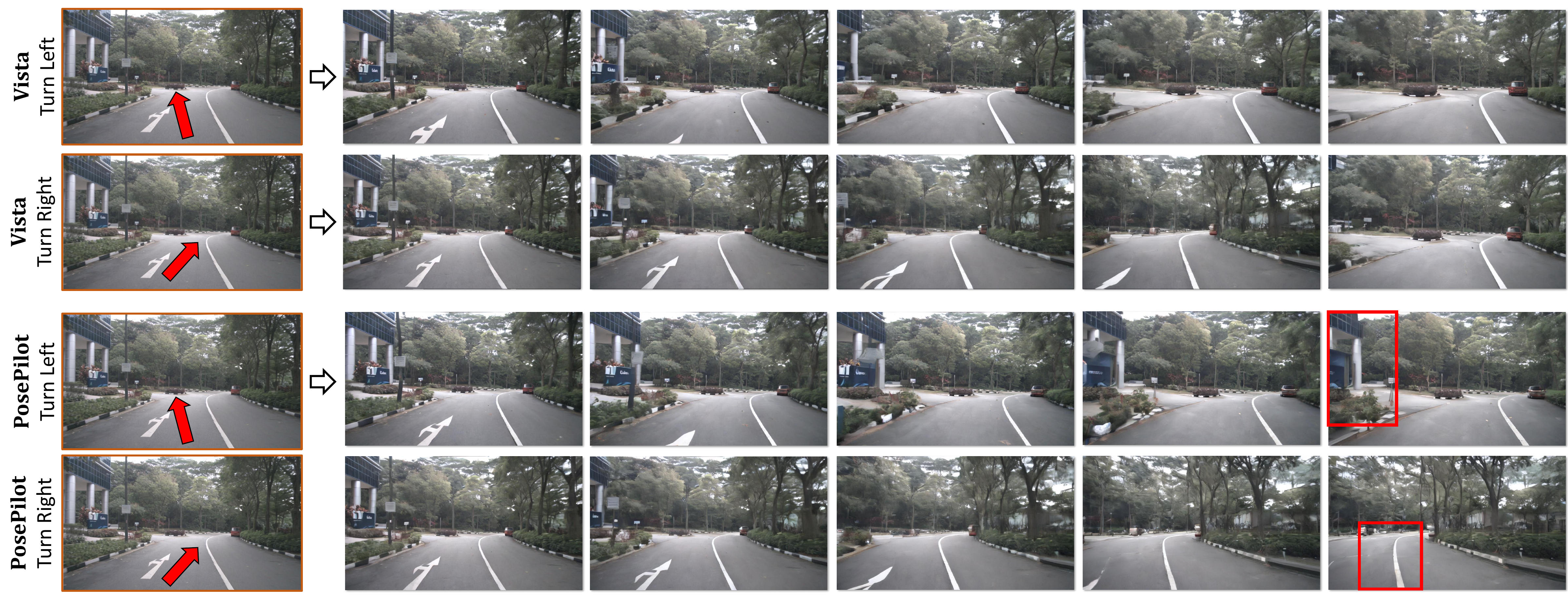}
    \vspace{-19pt}
    \caption{\textbf{Controllability of \method.} Our model can enhance the model's pose controllability, which generates videos that collaboratively align with the conditional inputs. More examples can be found in the supplementary video.}
    \label{fig:control}
\end{figure*}

\section{Experiment}
\subsection{Experiment Setup}
\noindent \textbf{Dataset.}
We adopt the nuScenes~\cite{caesar2020nuscenes} dataset for main experiments to evaluate the camera controllability of video generation in driving scenarios. We inherit the official split for evaluation, which includes 700 training videos and 150 validation videos, where each video sequence is recorded at 2 Hz and lasts about 20 seconds. Additionally, we follow~\cite{he2024cameractrl} to use the RealEstate10K~\cite{zhou2018stereo} dataset to conduct cross-domain application experiment in general video generation.

% \noindent \textbf{Training and inference.}
% We utilize AdamW optimizer \cite{loshchilov2017decoupled} and Cosine Annealing scheduler for training, where the learning rate is 1e-4 and the weight decay is 1e-2. The training is conducted on 8 NVIDIA A100 GPUs and lasts for 2 days.
% During the inference, we perform sampling inference using Rectified Flow \cite{liu2022flow} with 30 steps.   

\noindent \textbf{Evaluation Metrics.}
We adopt TranErr and RotErr \cite{he2024cameractrl} as camera alignment metrics to quantify the discrepancy between the translation and rotation vectors of the specified camera pose conditions and those of the generated video camera trajectories, assessing the quality of camera pose control during video generation. We also employ the Fréchet Inception Distance (FID) \cite{heusel2017gans}, along with parameters size, iteration time, and inference time, in ablation experiments to evaluate video quality, model size, and computational efficiency.

\subsection{Main Results}
We adapt our methods to the existing world model solutions on the nuScenes validation set.
We divide the nuScenes validation set into four subsets. The \texttt{TranErr} and \texttt{RotErr} scores are measured on each subset and then averaged. We conduct experiments on most of the publicly available world models, including diffusion-based methods \cite{jiang2024dive, gao2024vista} and auto-regression-based approach \cite{hu2024drivingworld}. The results are shown in Tab.~\ref{tab:main}.
We can see that with our \method, all of the models can get relatively lower \texttt{TranErr} \& \texttt{RotErr}, which demonstrates the effectiveness of the proposed architecture. For example, in diffusion-based methods, with our \method, Vista~\cite{gao2024vista} gets a 0.31 decrease on translation error and 0.21 decrease on rotation error. In auto-regression-based approach, Drivingworld~\cite{hu2024drivingworld} also gets a 0.22 decrease on translation error and 0.16 decrease on rotation error, indicating the superior generalization of our method.
We also show the qualitative results of \method~at Fig.~\ref{fig:control}, we can see that our \method~ contributes significantly to the model’s camera pose control. 

\begin{table*}[]
\begin{center} 
\caption{Element-wise Ablation Study}
\label{tab:ablation}
\begin{tabular}{@{}ccccccccc@{}}
\toprule
\multicolumn{3}{c}{\textbf{\method}} & \multirow{2}{*}{\textbf{\texttt{TransErr}}↓} & \multirow{2}{*}{\textbf{\texttt{RotErr}}↓} & \multirow{2}{*}{\textbf{FID}↓} & \multirow{2}{*}{\textbf{Parameters (M)}↓} & \multirow{2}{*}{\textbf{Iter Time (s)}↓} & \multirow{2}{*}{\textbf{Inference Time (s)}↓} \\ \cmidrule(r){1-3}
$L_{mse}$ & $L_{p}$ & $L_p^{\text{inv}}$ & & & & & & \\
\midrule
\ding{52} & \ding{52} & \ding{52} & \textbf{6.37} & \textbf{1.40} & \textbf{13.7} & +181.1 & 8.83 & 47.8 \\
\ding{52} & \ding{52} & & 7.56 & 1.67 & 14.5 & +90.5 & 8.71 & 47.7 \\
\ding{52} & & \ding{52} & 7.09 & 1.51 & 13.8 & +90.5 & 8.70 & 47.7 \\
 & \ding{52} & \ding{52} & 6.97 & 1.43 & 14.7 & +181.1 & 8.80 & 47.8 \\

\bottomrule
\end{tabular}
\end{center}
\vspace{-0.5cm}
\end{table*}

\subsection{Ablation Study}
To evaluate the impact of each proposed loss terms in \method, we conduct a comprehensive ablation study by analyzing their effects the key aspects of our architecture.

\noindent \textbf{Camera Pose Control Accuracy.}
We examine how each loss term contributes to precise camera pose control. By ablating specific losses, we assess potential deviations in predicted viewpoints and the overall stability of camera motion across frames. This is quantified using Translation Error (TransErr) and Rotation Error (RotErr) following common practice \cite{he2024cameractrl}. This analysis reveals that certain loss terms are crucial for maintaining low translation errors, ensuring accurate camera positioning along the intended path, while other losses are essential for preserving correct angular orientation and minimizing rotation errors. By identifying the specific contributions of each loss term, we can optimize the camera pose control system according to the unique demands of different scenarios, ultimately enhancing the performance and reliability of systems dependent on precise camera pose estimation.

% To evaluate the effectiveness of photometric loss in our \method~framework, we conduct experiments by removing this loss and comparing the results with the full model to assess how the absence of photometric loss impacts the overall learning process. 
% The results in Tab.~\ref{tab:ablation_photometric} show a significant degradation in performance without the photometric loss, particularly in the FID and SSIM. This demonstrates that the photometric loss plays a critical role in ensuring the network maintains high-quality reconstructions by penalizing inconsistencies between predicted and ground truth pixel intensities. Furthermore, we also provide qualitative results to show that the photometric loss aids in capturing fine-grained structural details, indicating its necessity for achieving structure-aware world models.

\noindent \textbf{Video Generation Quality.}
We evaluate how the removal of each loss affects the realism and consistency of the generated video. This includes assessing structural coherence, temporal smoothness, and fidelity to the intended viewpoint. The quality of video generation is measured using the Fréchet Inception Distance (FID) \cite{heusel2017gans}. We find that removing either photometric loss leads to a noticeable increase in FID, indicating degraded visual fidelity and weakened viewpoint alignment. In particular, the inverse loss contributes to improved temporal consistency, reducing flickering and motion discontinuities across frames. This highlights the role of geometric supervision not only in pose accuracy, but also in maintaining coherent, scene-consistent synthesis.

\noindent \textbf{Model Parameter Size.}
We analyze the influence of each loss term on the total parameter count of the model, determining whether certain losses introduce additional complexity or redundancy in the architecture. This is assessed through the parameter count of the model. Our findings indicate that, although the introduction of our method does increase parameters—specifically, an addition of 180 million parameters—this increase is accompanied by a substantial improvement in pose controllability. Importantly, even with the parameter increase, the size of our model remains considerably smaller than the original world models (more than 1 billion). This indicates a successful balance between maintaining a compact and efficient architecture while achieving greater capabilities in pose control. These insights suggest that carefully crafted loss terms can add value by refining model performance with minimal additional complexity, providing a strategic advantage in developing sophisticated yet efficient camera control systems.
% In previous experiments, we have demonstrated that the auxiliary depth estimation tasks and ego-motion prediction tasks can help improve the generation quality. One question is: will this significantly increase model latency or video memory usage? We measure the time taken for each training iteration consisting of forward and backward pass. In this experiment, we test the time with Tesla A100 GPU, and the training time is tested on 8 GPUs, and the inference time is tested on a single GPU. The resolution of the video is $1280 \times 720$ and the frame number is $8$.

\noindent \textbf{Iteration and Inference Efficiency.}
% We measure how different loss configurations impact training iteration speed and inference time, identifying potential trade-offs between computational efficiency and model performance. These are evaluated based on Iteration Time and Inference Time. \todo{To Be Add more}
We also measure how different loss configurations impact training iteration speed and inference time, identifying potential trade-offs between computational efficiency and model performance. In this experiment, we test the inference time with Tesla A100 GPU. The training time is tested on 8 GPUs and the inference time is tested on a single GPU. The resolution of the video is $1280 \times 720$ and the frame number is $8$. We can see that the iteration time and inference are generally the same, indicating the lightweight and plug-and-play ability of our \method.

\subsection{Cross-domain Application}
In previous experiments, we demonstrate that the introduced \method~significantly improves pose controllability in world models for autonomous driving. Surprisingly, when we apply this module to the general video generation task, it further enhances the precision of camera pose control, achieving a new state-of-the-art performance on RealEstate10K~\cite{zhou2018stereo}, as shown in Table~\ref{tab:realestate10k}. This finding indicates that our approach possesses strong cross-domain adaptability and scalability, paving the way for more advanced video generation tasks in diverse scenarios. Its reliance on self-supervised geometric signals makes it inherently adaptable across domains without task-specific tuning.

\begin{table}[]
\begin{center} 
\caption{Camera Control Ability in Natural Scenes.}
\label{tab:realestate10k}
\begin{tabular}{@{}lcc@{}}
\toprule
\textbf{Method} & \textbf{\texttt{TransErr}}↓ & \textbf{\texttt{RotErr}}↓ \\ 
\midrule
AnimateDiff~\cite{guo2023animatediff} & 9.81 & 1.03 \\
MotionCtrl~\cite{wang2024motionctrl} & 9.02 & 0.87 \\
CameraCtrl~\cite{he2024cameractrl} & 8.83 & 0.95 \\
\midrule
CameraCtrl + \method~\textbf{(Ours)} & \textbf{6.52} & \textbf{0.70}\\
\bottomrule
\end{tabular}
\end{center}
\vspace{-0.5cm}
\end{table}

\section{Conclusion}
In this paper, we presented \method, a lightweight and generalizable framework that significantly improves camera pose controllability in generative world models. By leveraging self-supervised structure-from-motion cues, \method~introduces a principled mechanism to couple depth and ego-motion estimation with video generation. Our approach enforces geometric consistency through pose-aware and inverse warping losses, while direct pose supervision further enhances viewpoint precision. Extensive experiments across autonomous driving and general-domain datasets show that \method~improves structural fidelity and motion alignment in diverse architectures. As a plug-and-play module, \method~requires no architectural changes and generalizes well across domains. We believe it offers a new perspective on integrating geometric priors into generative modeling, paving the way for controllable, physically grounded video synthesis.

% a lightweight yet powerful framework that significantly enhances camera pose controllability in generative world models. Drawing inspiration from self-supervised depth estimation, PosePilot leverages structure-from-motion principles to establish a tight coupling between camera pose and video generation. We incorporate self-supervised depth and pose readouts, allowing the model to infer depth and relative camera motion directly from video sequences. These outputs drive pose-aware frame warping, guided by a photometric warping loss that enforces geometric consistency across synthesized frames. 
% Extensive experiments on autonomous driving and general-domain video datasets demonstrate that PosePilot significantly enhances structural understanding and motion reasoning in both diffusion-based and auto-regressive world models. By steering camera pose with self-supervised depth, PosePilot sets a new benchmark for pose controllability, enabling physically consistent, reliable viewpoint synthesis in world models.

\newpage
\bibliographystyle{IEEEtran}
\bibliography{ref}

\color{red}
\end{document}